\documentclass[letterpaper, 10 pt, journal, twoside]{IEEEtran}

\IEEEoverridecommandlockouts                              

\usepackage{graphics} 
\usepackage{epsfig} 
\usepackage{mathptmx} 
\usepackage{times} 
\usepackage{amsmath} 
\usepackage{amssymb}  
\usepackage{mathtools}
\usepackage{tabularx, multirow}
\usepackage{color}
\usepackage{xcolor}
\usepackage{booktabs}
\usepackage{multirow}
\usepackage{url}

\usepackage{caption}
\begin{document}

\title{
Attentiveness Map Estimation for Haptic Teleoperation of Mobile Robot Obstacle Avoidance and Approach 
}

\author{Ninghan Zhong$^{1}$ and Kris Hauser$^{2}$
\thanks{Manuscript received: August 28, 2023; Revised: November 22, 2023; Accepted: December 25, 2023.}
\thanks{This paper was recommended for publication by Editor Jee-Hwan Ryu upon evaluation of the Associate Editor and Reviewers' comments.
This work was supported by NSF NRI Grant \#2024775.} 
\thanks{$^{1}$Ninghan Zhong is with the Department of Electrical and Computer Engineering, University of Waterloo, Canada. {\tt\footnotesize n5zhong@uwaterloo.ca}}
\thanks{$^{2}$Kris Hauser is with the Department of Computer Science, University of Illinois at Urbana-Champaign, USA. {\tt\footnotesize  kkhauser@illinois.edu}}%
\thanks{Digital Object Identifier (DOI): see top of this page.}
}


\markboth{IEEE Robotics and Automation Letters. Preprint Version. Accepted December, 2023} {Zhong \MakeLowercase{\textit{et al.}}: Attentiveness Map Estimation}

\maketitle

\begin{abstract}
Haptic feedback can improve safety of teleoperated robots when situational awareness is limited or operators are inattentive. Standard potential field approaches increase haptic resistance as an obstacle is approached, which is desirable when the operator is unaware of the obstacle but undesirable when the movement is intentional, such as when the operator wishes to inspect or manipulate an object. This paper presents a novel haptic teleoperation framework that estimates the operator's attentiveness to obstacles and dampens haptic feedback for intentional movement. A biologically-inspired attention model is developed based on computational working memory theories to integrate visual saliency estimation with spatial mapping. The attentiveness map is generated in real-time, and our system renders lower haptic forces for obstacles that the operator is estimated to be aware of. Experimental results in simulation show that the proposed framework outperforms haptic teleoperation without attentiveness estimation in terms of task performance, robot safety, and user experience.
\end{abstract}


\section{INTRODUCTION}
\IEEEPARstart{T}{Teleoperation} enables human operators to control mobile robots in complex environments in which human judgment and adaptability are necessary. However, operators may not be aware of obstacles, and proximity-based
haptic feedback is a popular approach to inform an operator of obstacles that pose a risk of collision~\cite{teleoperated_shared_control, emg_shared_control}.  Experiments have shown that haptic feedback is well-suited for alerting users to conditions that they are not visually aware of and require immediate response~\cite{haptic_ar_comparison}. A disadvantage of such feedback is that it can distract or annoy an operator who is alert and attentive to an obstacle. Moreover, for mobile manipulators that must approach or make contact with an obstacle, such as reaching for items on tables and shelves or pushing furniture, we do not wish for the robot to avoid contact, but rather {\em unintentional contact}.  For intentional movements toward an obstacle, repulsive haptic forces would ``fight'' against the operator, leading to control contention and frustration. Moreover, haptic proximity alerts such as vibrations would annoy and distract the operator. Experiments have shown that contentious haptic feedback both increases cognitive load and degrades task performance~\cite{neurobehavioral_assess, evaluation_haptic}. Recent studies have addressed this issue using intent prediction, such as predicting intent from a predefined task set~\cite{contextual_task_learning} or as a goal location~\cite{coffey_intent}. However, it is difficult to infer intentions in open-world scenarios where the set of possible tasks is hard to define and intent is ambiguous.

This paper introduces a new haptic teleoperation approach that provides haptic feedback to avoid unintentional approach toward obstacles, while allowing intentional approach toward obstacles to be unimpeded. The method is based on the premise that human attention is easier to model than intent, and attention-modulated haptic feedback is sufficient to avoid control contention during intentional obstacle approach. Our system is based on an attentiveness map estimation (AME) that continuously updates in a scalar field to approximate how attentive the human is to all the obstacles around the robot. Our biologically-inspired attention model is composed of {\em saliency estimation}~\cite{saliency_rapid_scene_analysis}, which estimates how likely is the operator to notice the obstacle at a point in time, and a {\em working memory model}~\cite{TBRS_2004_adult}, which estimates how likely the operator remembers obstacles currently being seen and previously seen.  It enables operators to intentionally approach obstacles while experiencing lower contention than standard potential field approaches and also provides haptic feedback to avoid obstacles that operators have not seen or may have forgotten. We reduce the haptic potential of an obstacle proportionally to its estimated attentiveness. The model does not require the environment to be known in advance, is updated in real-time, and is applicable to any mobile robot equipped an RGB-D camera.

\begin{figure*}[tb]
    \centering
    \includegraphics[width=0.95\linewidth]{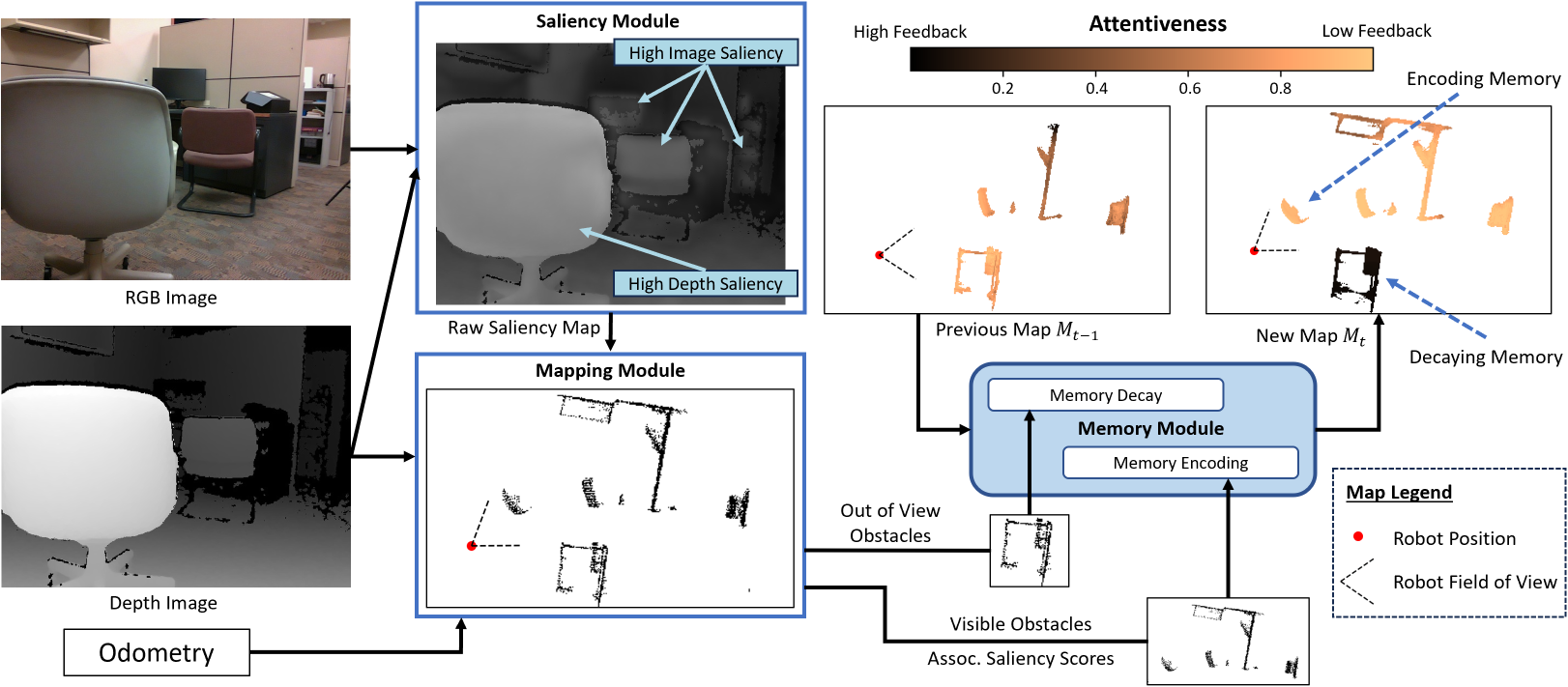}
    
    \caption{Diagram of the AME model.  {\em Saliency Module} uses the RGB-D image to generate a saliency map. {\em Mapping Module} computes a top-down occupancy map and pairs currently visible points with saliency scores. Visible obstacles enter {\em Memory Encoding} and out-of-view obstacles undergo {\em Memory Decay} in the {\em Memory Module}, which updates the attentiveness map. Timesteps are exaggerated for illustration.}
    \label{fig:ame_overview}
    \vspace{-15pt}
\end{figure*}

The method is evaluated in a human subjects study with $N=21$ subjects teleoperating a wheeled humanoid in simulated environments involving mixed obstacle avoidance and obstacle approach. Results demonstrate that the proposed framework outperforms haptic teleoperation without our attentiveness mapping approach. Specifically, operators complete tasks more efficiently, with fewer collisions, experience lower contention, and rate the system more favorably on subjective metrics. Additional examples are found in the supplemental video.

\section{Related work}
Haptic feedback is widely used for warning signals and shaping the user's control to aid in obstacle avoidance for mobile robots~\cite{GPF_original,Generating_artificial_force,GPF_Lam}. The most common strategy to avoid control contention during intentional obstacle approach is to predict the operator's intended task (e.g., a goal position) and shape haptic feedback accordingly. 
The techniques differ in how intent is predicted and how feedback is shaped.
Intent prediction is usually accomplished by modeling how the operator input signals map to one or more possible goals.  Gottardi \emph{et al.} \cite{teleop_intent_prediction} use maximum entropy inverse optimal control to predict the probability that a goal out of $N$ candidate goals is the next goal position. An artificial potential field (APF) is constructed such that robot is repelled from obstacles while being attracted toward the goals, where attraction scales with the predicted probabilities.  Gao and Zollner~\cite{contextual_task_learning} formulated intent as a finite set of tasks, which includes {\em Object Bypass} (i.e. obstacle avoidance) and {\em Object Inspection} (i.e. obstacle approach), and proposed a Gaussian Process classifier that takes as inputs the operator commands and environment information to predict the current task.  They did not, however, demonstrate their approach in an integrated haptic teleoperation system.

Finite-task approaches are limited by a dependence on prior knowledge of the environment, a known set of tasks, and/or teleoperation data collected in the deployed environment.  Other researchers have inferred the intended goal as a continuous variable, e.g., using the operator's velocity reference input and a look-ahead time~\cite{intent_with_input_VF,coffey_intent} to predict a goal position, or using a Hidden Markov Model to predict the intended movement direction and distance~\cite{aerial_intent_prediction}. In these prior works, an RRT-style planner is used to find an obstacle-free path toward the predicted goal, and the haptic feedback guides the operator along the path while avoiding obstacles. These continuous goal inference approaches cannot represent task ambiguity in the inferred goal, and this could lead to operator surprise and control contention when the inference is incorrect. We expect these issues to be most severe with inattentive operators, novice operators, movements toward obstacles that are outside of the operator's field of view, or exploratory movements that do not call for strong goal-seeking behavior. In contrast, our proposed model takes a perception-driven approach based on visual attention rather than the operator's input. We show that modeling what the operator sees and is attentive to allows our method to identify which obstacles are most important to avoid, and does not require prior assumptions or data about the environment or the tasks to perform.

Visual attention implies the focus of mental power upon certain objects or areas for more detailed observation and mental processing. Saliency prediction aims to model human visual attention as a saliency level across an image, where high saliency areas tend to attract more attention \cite{review_from_neuro_to_deep}. Classic saliency models can be classified into top-down and bottom-up. Top-down saliency models focus on task-dependent visual attention and require prior knowledge specific to the task and environment \cite{top_down_control}. In contrast, bottom-up methods focus on low-level visual stimuli such as color, contrast, and texture that naturally attract involuntary visual attention~\cite{saliency_rapid_scene_analysis,visual_attention_3D_video}. More recent studies leverage deep neural networks to predict visual attention by directly learning from eye-tracking datasets~\cite{RNN_video_saliency}. To ensure adaptability across environments and tasks, our work uses Itti's classical bottom-up saliency model \cite{saliency_rapid_scene_analysis}, which is one of the most influential models in the field~\cite{survey_of_most_recent} and does not rely on external training data. 

\section{Attentiveness Map Estimation}

The key component of our system is the Attentiveness Map Estimation (AME) model (Fig.~\ref{fig:ame_overview}).  AME continuously takes as input the RGB-D image viewed by the robot's camera, and estimates an attentiveness map that represents the operator's current attentiveness distribution across possible obstacles. It is comprised of saliency, mapping, and memory modules, which are described in detail here. Later, Section~\ref{sec:Haptic} describes how we use AME to modulate the haptic feedback. 

The goal of the model is to estimate in a scalar-valued map how likely the operator notices obstacles in their visual field and is aware of obstacles that were previously seen.  This requires estimating how visually distinctive obstacles are in the current image (saliency) as well as how the operator encodes viewed obstacles in their working memory.  The memory model assumes that attentiveness increases for salient obstacles and decays for obstacles outside of the visual field.  Our model uses biologically-inspired methods to implement these functions in a spatial mapping framework.

We note that saliency estimation is only a coarse approximation of attention and could be improved by additional instruments like eye tracking. Such eye gaze measurements could easily replace the saliency module, and we leave this option to future work.

\subsection{Saliency Module}

Our saliency module fuses bottom-up saliency detection on RGB and depth saliency detection on the depth image to produce a saliency map $S$, similar to Zhang \emph{et al.}~\cite{visual_attention_3D_video}.


We adopt Itti's bottom-up saliency model~\cite{saliency_rapid_scene_analysis} to obtain an image saliency map $S_{m}$ from the RGB image.   The claim is that visual neurons are sensitive to a small region (center) embedded inside a weaker and broader region (surround).  To estimate sensitivity computationally, low-level feature maps indicating intensity, color, and orientation are calculated at different scales and combined into a map $S_{m}$ assigning each pixel $(u, v)$ an image saliency score $S_{m}[u, v] \in [0, 1]$.  A Python implementation of the algorithm is used in our work\footnote{\url{https://github.com/akisatok/pySaliencyMap}}.

The depth saliency map assumes that closer regions receive higher saliency scores, similar to Zhang \emph{et al.}~\cite{visual_attention_3D_video}. Assume that $Z$ is the depth image and $Z[u, v]$ corresponds to the perceptive depth of pixel $(u, v)$, then the depth saliency score $S_{d}[u, v] \in [0, 1]$ for pixel $(u, v)$ is computed as:
\begin{equation}\label{eq:depth_saliency_map}
    S_{d}[u, v] =  \frac{z^n}{Z[u, v]} \cdot \frac{z^f - Z[u, v]}{z^f - z^n} 
\end{equation}
where $S_d$ is the depth saliency map and $z^f$ and $z^n$ are the farthest and closest depth, respectively. 

A final fusion step combines the image saliency map $S_{m}$ and the depth saliency map $S_d$ together to generate the final saliency map $S$. This is a linear combination of the image and depth saliency maps, $S = k_{m} \cdot S_{m} + k_d \cdot S_d$, 
where $k_{m}$ and $k_d$ are image and depth weighting factors that sum to 1, respectively. Fig.~\ref{fig:ame_overview} {\em Saliency Module} shows the combined effect of image and depth saliency detections.

\subsection{Mapping Module}
The mapping module reprojects points that are currently visible to the operator into a top-down view of the environment, and pairs these points with their associated saliency scores. 
Its inputs include the camera focal length $f_x, f_y$,  camera principal point $c_x, c_y$, camera transformation $[\mathbf{R}|\mathbf{t}]$ to the world frame (e.g., computed via an IMU, visual odometry, or SLAM system), the depth image, and the corresponding saliency map $S$. A pixel $(u, v)$ from the depth image with depth $z$ is reprojected to the world frame using:
\begin{equation}\label{eq:reprojection}
\begin{bmatrix}x_w \\ y_w \\ z_w \end{bmatrix} = 
\mathbf{R} 
\begin{bmatrix}
z/f_x & 0 & 0 \\
0 & z/f_y & 0  \\
0 & 0 & z  
\end{bmatrix}
\begin{bmatrix}u-c_x \\ v-c_y \\ 1 \end{bmatrix}
+\mathbf{t}.
\end{equation}
The point is also associated with saliency score $S[u,v]$.

Next, we discard points above the robot's height and below the ground plane, and drop the z-coordinate. The remaining points represent the visible x-y coordinates in the robot environment. They are then projected to a grid map $M$ of resolution $\beta$ to give a set of visible grid cells $P_v$. $\beta$ is a tunable parameter that can be adjusted based on the available computational power.

To calculate the saliency score of a grid cell, we take the minimum saliency score amongst all points that are projected to the cell. By taking the minimum saliency, we encourage the obstacle avoidance feedback to behave more conservatively, i.e., to respond strongly to an obstacle even if only a part of it has low salience. Overall, the module outputs a top-down saliency map $S_v$ which is defined for all currently visible grid cells.

\subsection{Memory Module}
The memory module continuously updates the estimated attentiveness map that represents the operator's current spatial attention distribution. To robustly model human attention, the memory module is formulated with a biologically-inspired design corresponding to the Time-Based Resource Sharing (TBRS) model \cite{TBRS_2004_adult}.

TBRS is a well-known model of working memory in cognitive neuroscience \cite{TBRS_math_transcription}.  Its main idea is that both encoding new memory and refreshing existing memory require mental focus\footnote{The term {\em mental focus} here refers to the concept of attention in the TBRS model. In the TBRS model, {\em attention} is conceptualized as a cognitive resource that could be focused to perform central processes. In the context of this paper, we use the terms {\em attention} and {\em attentiveness} interchangeably, which refer to the operator's spatial awareness of the robot surroundings.}, which is a limited mental resource. When mental focus is switched to encode new memory, existing memory suffers from a time-based decay. Further, there is a central bottleneck that allows only one central process. Consequently, at a time, either encoding new memory or refreshing existing memory can take place, but not both. Originally presented as a verbal theory, later studies \cite{TBRS_math_transcription, TBRS_computational_earliest} proposed computational versions of the TBRS model. 
We implement these versions using a memory encoding component that simulates the process of encoding new memories and a memory decay component that handles the forgetting of memories over time.

\subsubsection{From Saliency to Attention Rate}
High saliency areas in an image likely attract more visual attention \cite{review_from_neuro_to_deep}, but due to visual acuity being highest in the fovea of the eye, attention cannot be paid to all salient areas of the image simultaneously and instantaneously. To model the accumulation of attention over time, we map saliency to a temporal increase in attention, which we call the {\em attention rate}.
The attention rate $r_{x, y}$ for each visible grid point $p_{x, y} \in P_v$ is computed based on the associated saliency score $s_{x, y} \in S_v$.
We use a simple calculation that normalizes saliency scores across the visible grids:
\begin{equation}\label{eq:saliency_to_memory}
    r_{x, y} = \frac{s_{x, y}}{\sum_{S_v} s_{i, j}}.
\end{equation}
This suffices to incorporate three effects demonstrated from prior psychological studies: 1) visual saliency and working memory performance are positively correlated~\cite{visual_mem_positive2}, 2) high-saliency areas suppress the memory encoding of low-saliency areas \cite{high_exhaust_low}, and 3) the memory resource an area receives is relative to other areas' saliency in the scene \cite{relative_saliency}. In other words, salient areas receive more memory resources if they are presented among less salient areas.

\subsubsection{Memory Encoding}

Following the TBRS memory model from~\cite{TBRS_computational_earliest}, memory encoding is formulated as an exponential growth in $[0, 1]$ with 0 indicating no memory and 1 indicating full current attention. Let $m_{x, y}[t] \in [0, 1]$ be the attentiveness estimation for grid cell $p_{x, y} \in M$ at time $t$. For each visible grid cell $p_{x, y} \in P_v \subseteq M$, the attentiveness increases exponentially toward 1, with exponent proportional to the estimated attention rate:
\begin{equation}\label{eq:memory_encoding}
    m_{x, y}[t] = m_{x, y}[t-1] + r_{x, y} \cdot c \cdot (1 - m_{x, y}[t-1]).
\end{equation}
Here $c$ is a memory encoding scaling factor that defines the sensitivity of the memory encoding process. A smaller $c$ results in a more conservative model where the estimated operator awareness of the visible obstacles grows slower.

\subsubsection{Memory Decay}
TBRS models attentiveness in areas that are not currently visible as an exponential decay function. 
While \emph{memory encoding} updates visible grid cells, the decay component updates attention for non-visible cells $p_{x, y} \in M \setminus P_v$ according to:
\begin{equation}\label{eq:memory_decay}
    m_{x, y}[t] = (1-D)\cdot m_{x, y}[t-1].
\end{equation}
Here $D \in [0,1]$ is a constant decay rate that defines the sensitivity of the memory decay process. A  larger $D$ results in a more conservative model where the operator is estimated to forget recently seen obstacles faster.


We note that the TBRS model \cite{TBRS_2004_adult} provides an additional mechanism named \emph{memory refreshing} that models how humans might actively reflect on past experiences. This slows memory decay when the human is not looking at previously seen objects.
To be conservative for obstacle avoidance, we do not perform memory refreshing and assume that the operator's mental focus is always occupied with either the task at hand or other distractions in the visual field.

\subsection{Real-World Implementation}

\setlength{\belowcaptionskip}{-5pt}
\begin{figure}[tb]
    \centering
    \includegraphics[width=\linewidth]{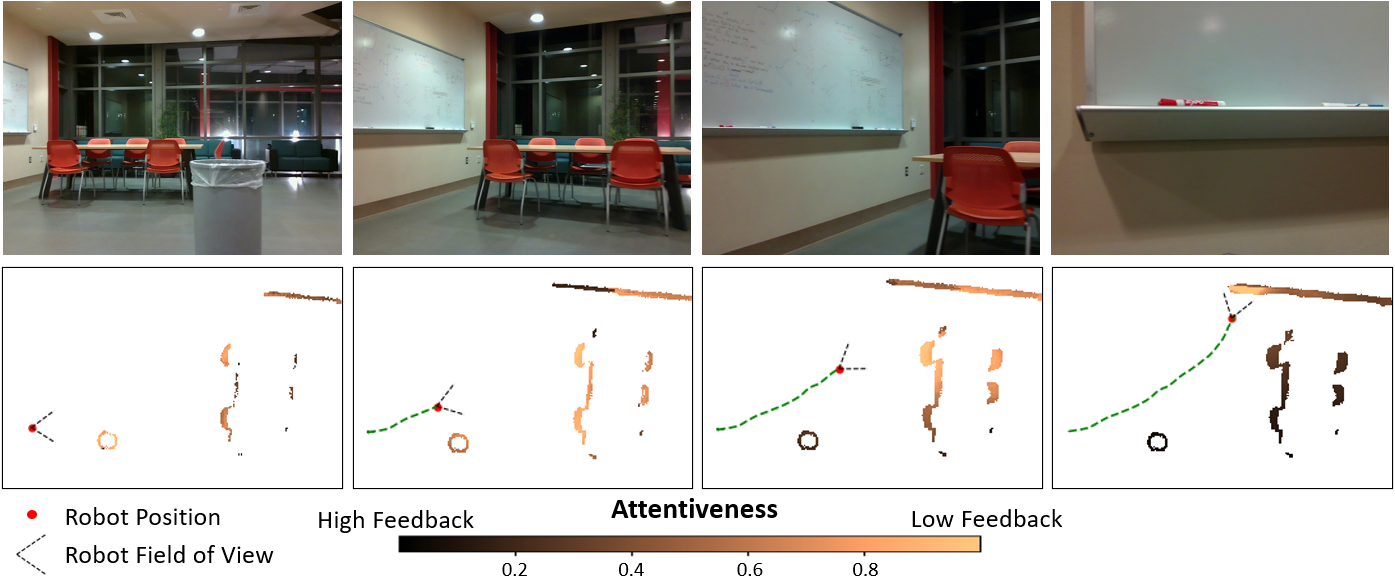}
    \caption{Attentiveness map updates while approaching a wall. Attentiveness increases for obstacles in view and decays for out-of-view obstacles. Brighter color represents obstacles with higher attentiveness, leading to a reduced repulsive force.}
    \label{fig:sequencial_fig}
\end{figure}
\setlength{\belowcaptionskip}{1ex}

\label{sec:real_world}
We implemented AME on a perception system consisting of an Intel RealSense L515 camera and an Intel RealSense T265 camera. The L515 camera provides RGB-D frames at 30\,Hz, and the T265 camera provides odometry estimations at 30\,Hz from its built-in tracking system.  We use RTAB-MAP \cite{rtabmap} on an Intel NUC 12 Pro with a 10\,Hz detection rate to generate an occupancy map and the pose estimates. AME runs at 10\,Hz, and both occupancy map and attentiveness map have a resolution of 2\,cm. Parameters of the algorithm are set to $c=50$, $D=0.04$, and $k_m=k_d=0.5$ as in the rest of our experiments.
 Fig.~\ref{fig:sequencial_fig} provides a working illustration of obstacle attentiveness updates from the AME. 

\section{HAPTIC FEEDBACK}
\label{sec:Haptic}

\setlength{\belowcaptionskip}{-10pt}
\begin{figure}[t]
    \centering
    \includegraphics[width=\linewidth]{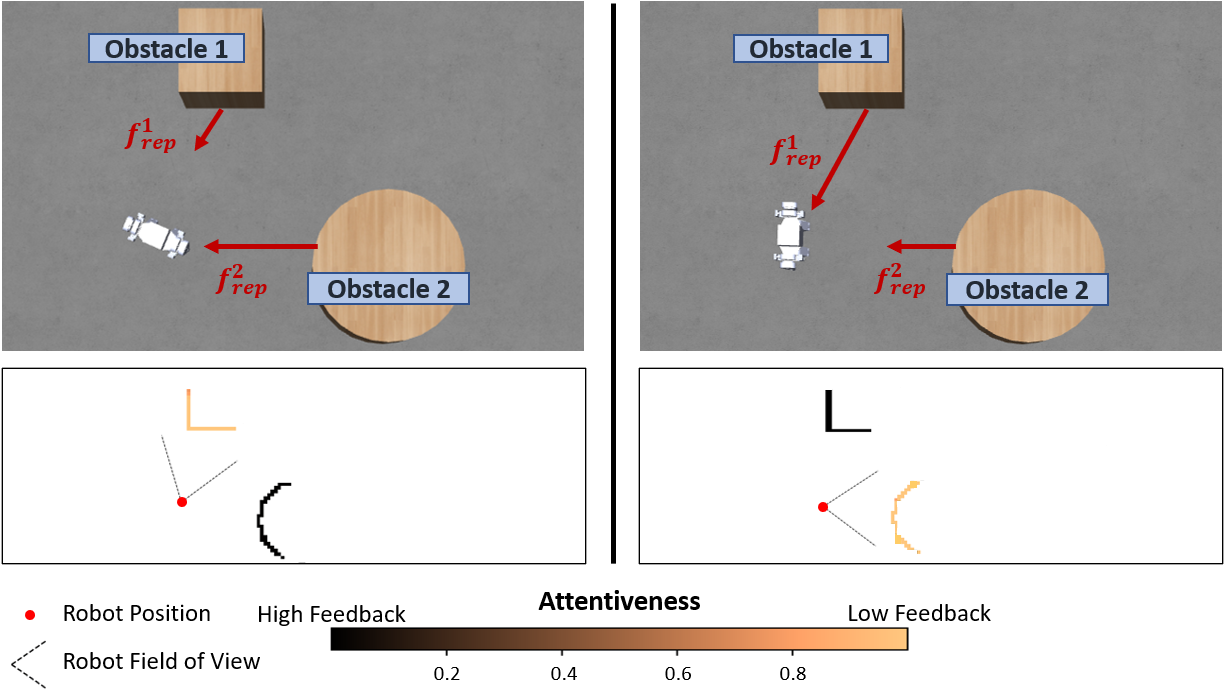}
    \caption{Estimated attentiveness modulates haptic feedback.   Left: the operator's attentiveness is high around obstacle 1, resulting in a stronger obstacle 2 repulsion $f_{rep}^{2}$ due to lack of awareness on obstacle 2. Right: the operator's attentiveness is high around obstacle 2, resulting in a stronger obstacle 1 repulsion $f_{rep}^{1}$.}
    \label{fig:haptic_visual}
\end{figure}
\setlength{\belowcaptionskip}{1ex}

Our haptic feedback module reduces the repulsive force generated from obstacles that the operator is attentive to, which has the effect that less repulsion is generated when approaching an obstacle directly from the line of sight of the operator, whereas full feedback is provided when the operator backs into out-of-view obstacles or steers into obstacles that suddenly appear in view. 
Our Attention-Modulated Generalized Potential Field (AMGPF) extends the Generalized Potential Field (GPF) method \cite{GPF_original}, which jointly considers obstacle distance and relative velocity.  
Following GPF, for each obstacle $o_i$, AMGPF computes reserve avoidance time:
\begin{equation}\label{eq:reserve_time}
    t_{res}(d_i, v_i) = \frac{d_i}{v_i}
\end{equation}
where $d_i$ is the robot's distance to $o_i$ and $v_i$ is the relative velocity in the direction of $o_i$. For $t_{safe}$ as a safe reserve time threshold, and $d_{safe}$ as a safe distance threshold, the combined risk factor for the obstacle is:
\begin{equation}\label{eq:risk_factor}
    r(d_i, v_i) = r_t(d_i, v_i) + \alpha \cdot r_d(d_i)
\end{equation}
where $r_t(d_i, v_i) = \frac{1}{t_{res}(d_i, v_i)} - \frac{1}{t_{safe}}$ is the temporal risk factor and $r_d(d_i) = \frac{1}{d_i} - \frac{1}{d_{safe}}$ is the distance risk factor, and are both lower-bounded by 0. $\alpha$ is a scalar that controls the balance between temporal and distance risks. The non-modulated repulsion from obstacle $o_i$ is computed as:
\begin{equation}\label{eq:repulsion_obstacle}
    R(d_i, v_i) = \begin{cases} 
      1 & r(d_i, v_i) \geq \frac{1}{G} \\
      G \cdot r(d_i, v_i) & \text{otherwise}
   \end{cases}
\end{equation}
where $G$ is a positive gain value to adjust field sensitivity. In the presence of multiple obstacles, a combined repulsion is generated by considering the risk from each obstacle. 

To incorporate attention, we assume that an obstacle with a high attention level is less likely to pose risk than an obstacle with an equal risk factor but with a low attention level. Specifically, assume obstacle $o_i$ is at position $(x_i, y_i)$, and the estimated attentiveness at the obstacle position is $m_{x_i, y_i}$, the obstacle repulsion with attentiveness estimation is:
\begin{equation}\label{eq:repulsion_obstacle_attn}
    R_{attn}(d_i, v_i, x_i, y_i) = R(d_i, v_i) \cdot (1 - \gamma \cdot m_{x_i, y_i})
\end{equation}
where $\gamma$ is a tunable parameter in range $(0, 1]$ that dictates how much to reduce repulsion based on obstacle attention. This allows the system to render a light warning force even when attention to an obstacle is full (i.e. $m_{x_i, y_i} = 1$) or when the attention is overestimated. Fig.~\ref{fig:haptic_visual} provides an illustration of how estimated attentiveness modulates the haptic feedback from each obstacle.  
The total repulsion from all obstacles $\mathbf{O}$ is then computed:
\begin{equation}\label{eq:repulsion_total_attn}
    f_{total} = \sum_{(x_i, y_i) \in \mathbf{O}} w_i \cdot \hat{\mathbf{t}}_i \cdot  R_{attn}(d_i, v_i, x_i, y_i)
\end{equation}
where $\hat{\mathbf{t}}_i$ is the unit vector from $o_i$ to the robot and $w_i$ is a weight term.  With a uniform weight we found that the repulsion from small but ``risky'' obstacles could be canceled by forces from large, less risky obstacles. Instead, our implementation boosts the effect of high-risk obstacles using the nonuniform weight:
\begin{equation}\label{eq:repulsion_weights}
    w_i = \frac{R_{attn}(d_i, v_i, x_i, y_i)^n}{\sum_{(x_j, y_j) \in \mathbf{O}} R_{attn}(d_j, v_j, x_j, y_j)^n}
\end{equation}
where $n$ is a tunable risk emphasis factor. A uniform weight corresponds to $n=0$, while in our setup, we chose $n=1$.

\subsection{Implementation and Illustration}

\setlength{\belowcaptionskip}{-10pt}
\begin{figure}[tbp]
    \centering
    \includegraphics[width=0.9\linewidth]{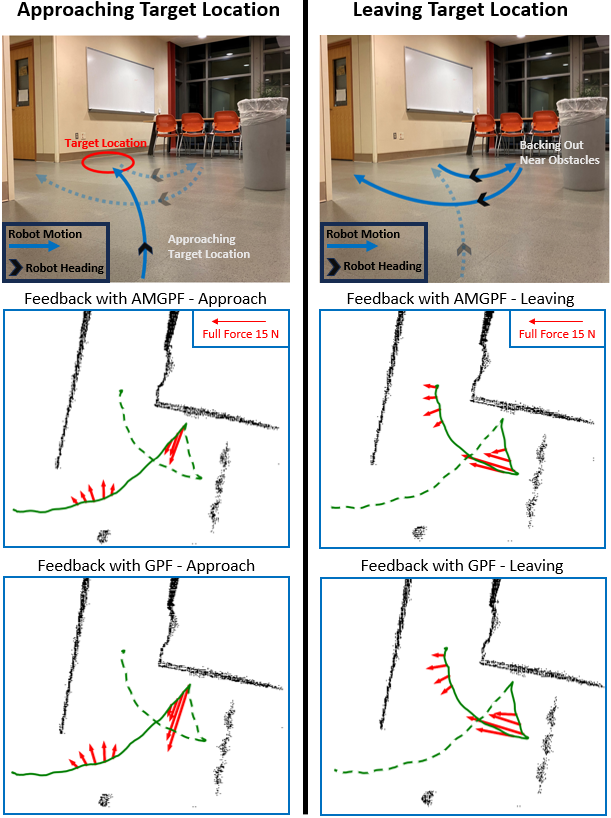}
    \caption{Haptic feedback forces, indicated by red arrows, of AMGPF (ours) vs standard GPF along an evaluation trajectory. The trajectory is split into approach (left) and leaving (right) segments. The higher estimated attentiveness to the target location during approach causes AMGPF to reduce repulsive forces. (Best viewed in color)}
    \label{fig:real_world_prelim}
\end{figure}
\setlength{\belowcaptionskip}{1ex}

%
Based on preliminary rounds of testing and informal feedback from the operators, $\gamma=0.65$ is chosen. Fig.~\ref{fig:real_world_prelim} illustrates the effect of AMGPF compared to GPF.
To compare the methods, we use the same recorded trajectory and plot the feedback forces that would have been generated. In the approach segment (Fig.~\ref{fig:real_world_prelim}, left), the robot passes by a trash bin and approaches the target whiteboard. GPF generates strong repulsive feedback when approaching the target, while AMGPF generates about half as much feedback, estimating that the operator is attentive during approach. The feedback does not drop down to zero, as specified by the parameter $\gamma$. In the leaving segment (Fig.~\ref{fig:real_world_prelim}, right) the robot backs away from the target, running a risk of colliding with the chairs that are out-of-sight of the camera. Both AMGPF and GPF render strong haptic feedback during backing up to assist collision avoidance and moderate feedback to steer around the corner. 

\vspace{2mm}
\section{Human Subject Evaluation}
\label{sec:Evaluation}

\setlength{\belowcaptionskip}{-10pt}
\begin{figure}[t]
    \centering
    \includegraphics[width=\linewidth]{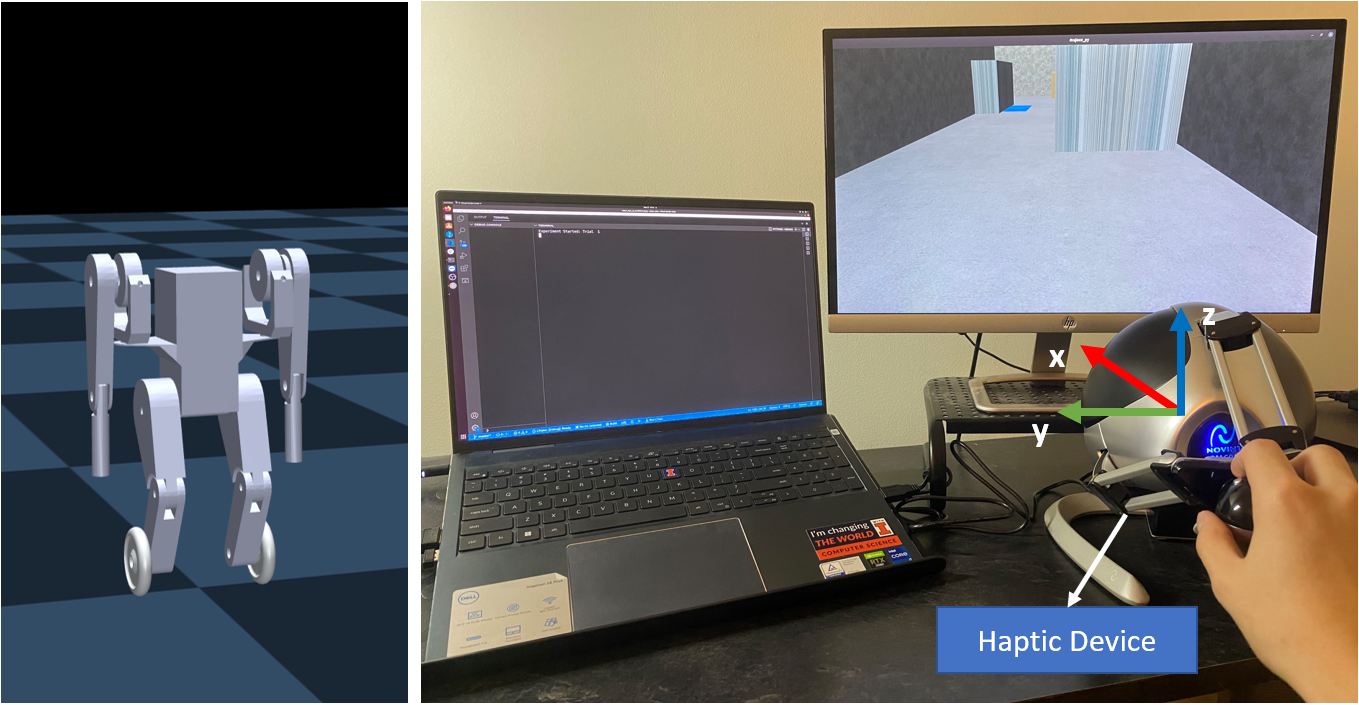}
    \caption{Simulated SATYRR robot and experiment setup
    }
    \label{fig:exp_setup}
\end{figure}
\setlength{\belowcaptionskip}{1ex}

\begin{table*}[tbp]
\caption{Experimental results ($N=21$) comparing AMGPF, GPF, and LH. The significance of the results is indicated by ``***'', ``**'', and ''*`` if respectively $p \le 0.001$, $p \le 0.01$, and $p \le 0.05$. $\uparrow$ indicates higher is better; $\downarrow$ indicates lower is better.}
\vspace{-5pt}
\label{table:exp_results}
\begin{center}
\setlength\tabcolsep{3pt}
\begin{tabular}{lccccccc}
    \toprule
      & Models & Completion time$\downarrow$ (s) & Collisions$\downarrow$ & Total distance$\downarrow$ (m) & Average speed$\uparrow$ (m/s) & TLX$\downarrow$ & Average force$\downarrow$ (N) \\
    \midrule
    Average & AMGPF (Ours)  & $162.3 \pm 4.9$ & $0.56 \pm 0.12$ & $54.11 \pm 0.82$ & $0.372 \pm 0.011$ & $36.78 \pm 1.86$ & $5.11 \pm 0.00$ \\
    & GPF  & $181.8 \pm 5.9$ & $0.67 \pm 0.18$ & $55.82 \pm 1.03$ & $0.342 \pm 0.010$ & $45.49 \pm 1.87$ & $6.93 \pm 0.01$\\
    & LH & $167.9 \pm 6.4$ & $1.81 \pm 0.31$ & $53.54 \pm 0.69$ & $0.366 \pm 0.012$ & $34.66 \pm 1.61$ & $2.40 \pm 0.00$ \\

    \cmidrule(l){2-8}
    ANOVA $F$-Score &  & 12.5449 & 9.5770 & 2.8758 & 3.1802 & 20.0867 & 464.1455 \\

    \cmidrule(l){2-8}
    ANOVA $p$-value &  & 0.0001$^{***}$ & 0.0004$^{***}$ & 0.0681 & 0.0523 & $<0.0001^{***}$ & $<0.0001^{***}$ \\
    
    \cmidrule(l){2-8}
    Post-hoc $p$-value & AMGPF v. GPF & 0.0433$^{*}$ & 0.9339 & - & - & 0.0018$^{**}$ & $<0.0001^{***}$ \\
    & AMGPF v. LH & 0.7635 & 0.0002$^{***}$ & - & - & 0.6792 & $<0.0001^{***}$\\
    & GPF v. LH & 0.2084 & 0.0008$^{***}$ & - & - & 0.0001$^{***}$ & 
    $<0.0001^{***}$ \\
    \bottomrule
\end{tabular}
\end{center}
\vspace{-15pt}
\end{table*}

\subsection{Experiment Setup}
We conduct experiments on a simulated wheeled humanoid robot performing navigation and mock manipulation tasks.  The platform is the wheeled humanoid robot SATYRR (Fig.~\ref{fig:exp_setup}, left)~\cite{satyrr_info_new}, and the robot's dynamics and interactions with obstacles are simulated in the MuJoCo physics simulator.
Subjects observe a simulated first-person view from the robot, and another monitor displays instructions (Fig.~\ref{fig:exp_setup}, right).
The robot is operated via a haptic control device that provides 3-DOF force feedback (Novint Falcon). 

A vertical z-axis damping force feedback is constantly applied to restrict the operator to manipulate the haptic device in the x-y plane. The device's position along the x-axis and the y-axis are mapped to the target forward velocity and angular velocity for the robot, respectively. To help the operators stabilize the robot, a light damping force toward the zero position is constantly applied. We incorporated a deadband to prevent unwanted robot movements from unintentional small displacements, and outside of the deadband the position is mapped linearly to the desired velocity. The target velocities are fed into a Linear Quadratic Regulator (LQR) controller to determine wheel torques, and a wheeled inverted pendulum (WIP) system is used to stabilize the robot's pitch.

\setlength{\belowcaptionskip}{-15pt}
\begin{figure}[!t]
    \centering
    \includegraphics[width=\linewidth]{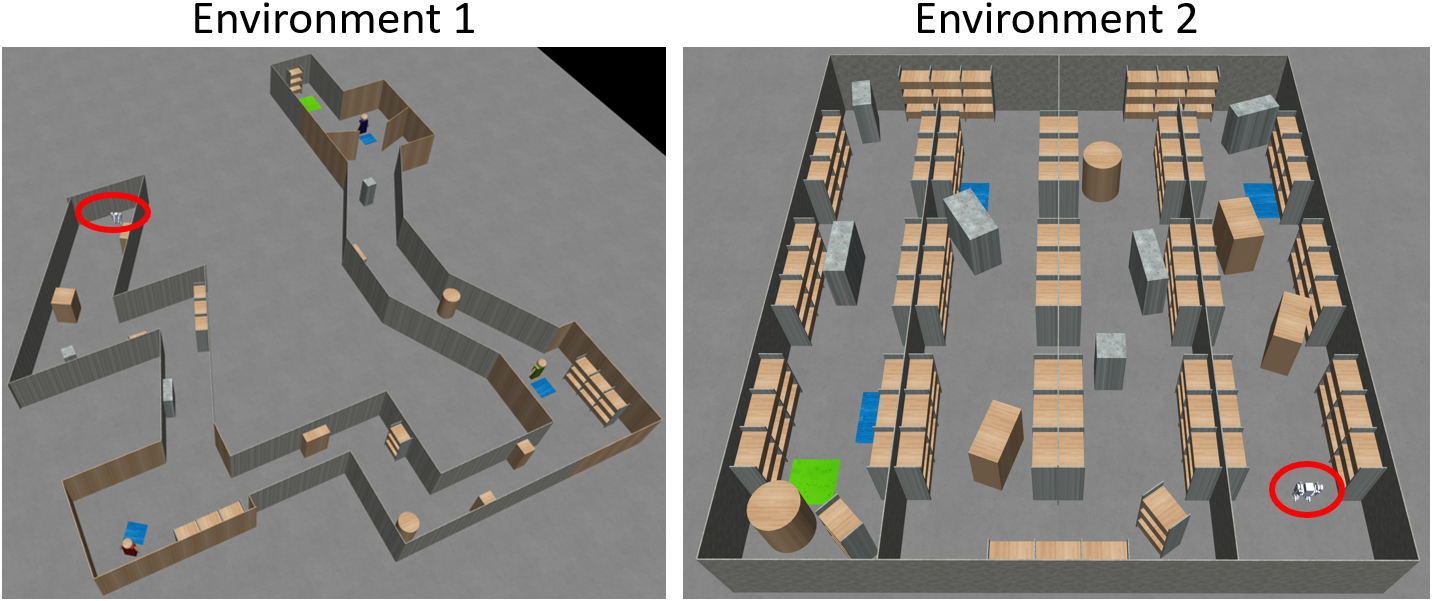}
    \caption{The two virtual environments used for evaluation. From the robot's initial position (circled in red) it must reach the green goal region after approaching each working area (blue) and remaining inside for 15\,s to perform a simulated task. (Best viewed in color)}
    \label{fig:envs}
\end{figure}
\setlength{\belowcaptionskip}{1ex}


\subsection{Experimental Procedures}

We compare the proposed attention-modulated haptic system (AMGPF) against two baseline systems. The first baseline provides haptic feedback but does not modulate feedback strength using attentiveness estimates (``vanilla'' GPF).  The second baseline, Low-Haptic (LH), is a system that does not render haptic feedback for obstacle avoidance, but only the light spring damping force toward the device zero position. 

For each trial, the subjects are asked to control the robot to approach each of three blue {\em working areas} and reach a green {\em goal region}, as shown in Fig.~\ref{fig:envs}. At each of the working areas, the robot is next to an obstacle, and the robot must remain inside continuously for 15\,s to simulate some kind of task. When the robot is in the middle of the working areas, we find that vanilla GPF produces approximately 10\,N of haptic feedback. Environment 1 (E1) simulates an indoor hallway and the working areas represent workstations for manipulation tasks. Environment 2 (E2) simulates a warehouse and the working areas represent shelves for pick-and-place tasks. E2 is designed to be more challenging than E1 with denser obstacles and less pathway clearance.  Once tasks are completed, driving the robot into the goal region completes the trial. The subjects are instructed to complete each trial as fast as possible while avoiding collisions. Both environments are linear to prevent users from taking alternative paths that might affect experimental results.

Subjects are informed that they will be using three different haptic feedback systems labeled System A, System B, and System C with letters assigned at random to the three experimental conditions.  Each subject is first trained to operate the robot in an obstacle-free environment to get familiar with the basic controls for 10 minutes. The subject then runs three mock trials, one for each system, in a training environment with obstacles to get familiar with trial tasks. Then, each subject runs each haptic system on each environment twice for a total of 12 recorded trials. The 12 trials are completed in random order under the constraint that environments and haptic systems are non-repeating. This is to minimize and balance learning effects on both haptic systems and environments. Before each trial, the subject is informed which haptic system they are about to use so that they can hopefully remember how the robot behaved under that system during training. After each trial, the subjects complete a NASA-TLX questionnaire to rate the trial experience.

\subsection{Evaluation Metrics}
Following Ju and Son \cite{evaluation_haptic}, trials are evaluated along three aspects: task performance, robot safety/stability, and control effort. Task performance consists of three metrics: trial completion time (s), total distance traveled (m), and average speed (m/s). The robot safety/stability metric is the total number of collisions per trial. The control effort metrics consist of the average feedback force (N) rendered to the operator and the task load index (TLX) assessed by the NASA-TLX questionnaire, which ranges from 0 to 100. 

\section{Results and Discussions}
\label{sec:Results}


The experiment was approved by the Institutional Review Board of UIUC with an exempt determination (IRB \#23559). A total of 21 participants were selected for the experiment. Evaluation results are summarized in Table~\ref{table:exp_results}. The evaluation metrics were first analyzed through repeated measures ANOVA. Post-hoc analysis was conducted to make pairwise model comparisons on significant ANOVA results. A p-value less than $0.05$ indicates a statistically significant difference.

Completion time using AMGPF is significantly lower than that of GPF and indistinguishable from LH. This suggests that operators are more efficient with less interference from force feedback during intentional obstacle approach. Similar observations can be made from total distance and average speed, but these differences were not found to be statistically significant. 
Furthermore, repulsive haptic feedback in AMGPF and GPF significantly reduces the collision rate compared to LH, which suggests that AMGPF assists with unintentional obstacle avoidance.  AMGPF has a slightly lower collision rate than GPF, which could be due to less control contention, but this difference is not statistically significant. 

The participants rate AMGPF and LH as having significantly lower Task Load Index than GPF, and no significant difference was found between AMGPF and LH. This also suggests that operators did not experience significant control contention with AMGPF. Corroborating this interpretation, we find that the average feedback force for AMGPF was lower than than of GPF. However, the feedback force was larger than that of LH.  Overall, these results indicate that AMGPF inherits the safety benefits from GPF while avoiding control contention to a large extent.

\section{CONCLUSION}
In this paper, a haptic control framework with a novel attentiveness map estimation (AME) model is presented. AME is a biologically inspired estimator of the operator's spatial attentiveness that does not rely on prior training data or information about the environment. By reducing repulsive forces from obstacles with high attentiveness estimates, the haptic control system was shown to be more user-friendly and effective in simulated tasks involving combined obstacle avoidance and obstacle approach.

These results raise several interesting questions for future work. Our current system assumes that attention is driven by the entire image, and does not incorporate eye gaze information, which is likely to be more informative than pure images.  Moreover, we do not estimate the operator's intended task, or whether operators are alert or distracted, and better signals for estimating these mental states is also likely to improve attentiveness estimation accuracy.

\section{Acknowledgement}
This work is supported by NSF National Robotics Initiative Grant \#2024775. We appreciate UIUC Intelligent Motion Lab members for their helpful comments and suggestions.

\bibliographystyle{IEEEtran}
\bibliography{bib}

\end{document}